%% file: main.tex
\documentclass{article}

\usepackage[numbers]{natbib} 
\PassOptionsToPackage{numbers, compress}{natbib}
\usepackage[preprint]{neurips_2024}

\usepackage[utf8]{inputenc} 
\usepackage[T1]{fontenc}    
\usepackage{hyperref}       
\usepackage{url}            
\usepackage{booktabs}       
\usepackage{amsfonts}       
\usepackage{nicefrac}       
\usepackage{microtype}      
\usepackage{xcolor}         
\usepackage{amsmath}
\usepackage{graphicx}
\usepackage{pgfplots}
\usepgfplotslibrary{external}
\definecolor{golden}{RGB}{255, 215, 0}
\usepackage{color, colortbl}
\usepackage{multirow}

\title{Conceptual Learning via Embedding Approximations for Reinforcing Interpretability and Transparency}

\author{%
   Maor Dikter\,\,, Tsachi Blau\,\,, Chaim Baskin \\
   Technion – Israeli Institute of Technology\\
 \texttt{\{maor.dikter,tsachiblau\}@campus.technion.ac.il}
 \\
 \texttt{\{chaimbaskin\}@technion.ac.il}
}

\begin{document}

\maketitle

\input{Sec/0_abstract}

\input{Sec/1_introduction}

\input{Sec/2_related-work}

\input{Sec/3_method}

\input{Sec/4_experiments}

\input{Sec/5_limitation}

\input{Sec/6_conclusion}

\bibliographystyle{plainnat}
\bibliography{Styles/bib}


\input{Sec/8_appendix}

\end{document}

%% file: Sec/0_abstract.tex
\begin{abstract}
\label{sec:abstract}

Concept bottleneck models (CBMs) have emerged as critical tools in domains where interpretability is paramount.
These models rely on predefined textual descriptions, referred to as concepts, to inform their decision-making process and offer more accurate reasoning.
As a result, the selection of concepts used in the model is of utmost significance.
This study proposes
\underline{\textbf{C}}onceptual \underline{\textbf{L}}earning via \underline{\textbf{E}}mbedding \underline{\textbf{A}}pproximations for \underline{\textbf{R}}einforcing Interpretability and Transparency, abbreviated as CLEAR, a framework for constructing a CBM for image classification.
Using score matching and Langevin sampling, we approximate the embedding of concepts within the latent space of a vision-language model (VLM) by learning the scores associated with the joint distribution of images and concepts.
A concept selection process is then employed to optimize the similarity between the learned embeddings and the predefined ones.
The derived bottleneck offers insights into the CBM's decision-making process, enabling more comprehensive interpretations.
Our approach was evaluated through extensive experiments and achieved state-of-the-art performance on various benchmarks. The code for our experiments is available at \href{https://github.com/clearProject/CLEAR/tree/main}{https://github.com/clearProject/CLEAR/tree/main}.

\end{abstract}

%% file: Sec/1_introduction.tex
\section{Introduction}
\label{sec:intro}

The unprecedented increase in the utilization of neural networks across diverse fields 
has highlighted the need for broader insight into how decisions are made.
As deep networks grow in complexity, focusing on explaining the decision using a post-hoc method~\cite{LRP, SHAP, tan2018distill}, which performs an analysis of the model after its training, is the best path to understanding.
Nevertheless, in sensitive fields such as healthcare, relying solely on such explanations is insufficient, and understanding the elements that shape the decision and the reasoning behind it is essential. 

To address the need for a thorough understanding of the inner workings of models, interpretable-by-design models~\cite{rudin2019stop} are being used. These inherently interpretable models integrate explanations within their architecture, thereby ensuring transparency and offering a more reliable form of reasoning.
Concept bottleneck models (CBMs)~\cite{koh2020concept} are one type of interpretable-by-design models.
The idea underlying such models is, given an input, first predict an intermediate set of specified concepts and then use this set to predict the target directly.
These models enable interpretation in terms of high-level concepts and allow for human interaction.
Like many contemporary challenges, these frameworks often employ multi-modal models. 
Aiming to quantify the relationship between an image and its corresponding textual representations, leveraging VLMs such as CLIP~\cite{CLIP} becomes a natural choice. 
The embedding space of these models enables us to gauge 
their alignment.

Today, methods for identifying concepts in the bottleneck commonly rely on Large Language Models (LLMs), for example, GPT-3~\cite{GPT-3}.
These techniques revolve around employing diverse prompts to guide the LLMs in understanding and extracting meaningful concepts. 
A recent work by Yang et al.~\cite{yang2023language} proposed LaBo, a framework for constructing a CBM. By generating descriptions and employing submodular optimization, LaBo selects relevant concepts for each class.
The framework then uses cosine similarity between the encoded image and concepts to make its predictions. 
Following this path, \citet{yan2023learning} demonstrate that relevant concepts can be derived from an approximation of the embedding space of VLMs, which then enable the textual descriptions to be located using nearest neighbor search. They anchored the learning of concepts using the Mahalanobis distance~\cite{mahalanobis2018generalized}, a statistical measure that evaluates the distance of a point from a distribution.

While these methods present notable strengths, they also include certain limitations that merit consideration. LaBo~\cite{yang2023language}, for example, requires extensive annotations to accurately represent the data. An overly expansive bottleneck is incomprehensible to humans, could compromise the quality of concepts, and, as demonstrated in \citet{yan2023learning}, often achieves comparable results to the use of random annotations.
On the other hand, while it is essential to formulate a small and accurate set of descriptors that efficiently represents our data, achieving so with a rough estimation from a limited number of descriptors falls short. In \citet{yan2023learning}, the set of concepts is relatively small, which limits the framework's ability to achieve a comprehensive representation.
A relatively large set of prior concepts is necessary to accurately estimate the distribution of textual embeddings. 
Once this distribution is established, reducing the number of concepts can improve interpretability. 
Interestingly, their approach sometimes showed improved performance without using the Mahalanobis distance, suggesting that the embedding space representation of the VLM may be suboptimal.
Furthermore, the method's concept selection process, being greedy, may lead to less-than-optimal outcomes.

Our approach addresses the aforementioned limitations effectively, as presented in Figure~\ref{fig:teaser}, which illustrates the main ideas of our framework.
To gain accurate approximations, we use a large pool of prior concepts, a descriptor pool. 
This strategy enables the learning of a set of embeddings for our textual concepts by training them as a single linear layer within our model.
This training process dynamically refines the embeddings based on the model's learning, ensuring they are fine-tuned to represent the concepts accurately.
To guide the learning of concepts appropriately toward meaningful embeddings, we use Langevin sampling to generate concept approximations from the probability density function (PDF) of the joint distribution of images and descriptors within the VLM's embedding space. Estimating a PDF can be challenging due to an intractable normalizing constant. In score matching, the density of the data is estimated by learning the gradient of the data distribution’s log-likelihood. Langevin dynamics then provides a method for sampling from a distribution using only its score function. 
By obtaining the score function, our technique directs our learned embeddings toward areas of higher density within the joint image–concept distribution—crucial for effective concept learning.
Following this, we construct a similarity matrix by calculating the cosine similarity between the learned embeddings and the descriptor pool. 
Subsequently, employing the Hungarian method~\cite{kuhn1955hungarian}, we establish a maximum perfect matching, which identifies the optimal set of textual representations based on their estimations. This approach ensures that our model's embeddings are both accurate and representative of the underlying conceptual structure.
\input{figures/teaser}

Overall, our framework offers three main contributions:

\begin{itemize}
  \item We model the joint distribution density of images and concepts using a score-matching based method, which enables us to develop a novel approach for concept embedding learning via sampling from this distribution.
  \item We introduce a concept selection methodology that achieves optimal allocation by optimizing joint similarity and develop an interpretable, data-adaptive CBM that surpasses existing models in performance. 
  \item We provide a comprehensive analysis of our framework's components and demonstrate its interpretability capabilities.
\end{itemize}

%% file: figures/teaser.tex
\begin{figure*}[t]
    \centering
    \includegraphics[width=\textwidth]{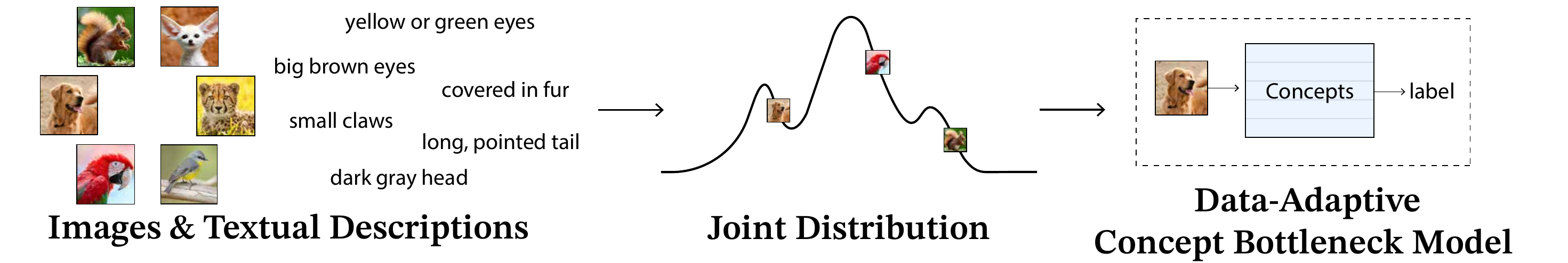}
    \caption{The core components of CLEAR, our proposed paradigm for constructing a data-adaptive CBM by modeling the joint distribution of images and concepts.}
    \label{fig:teaser}
\end{figure*}

%% file: Sec/2_related-work.tex
\section{Related work}
\label{sec:related-work}

Explainability methods have gained prominence for their utility in demystifying model decisions. These methods often employ saliency-based explanations to highlight input features that significantly influence predictions, with common techniques including feature importance scores~\cite{LRP, SHAP, ribeiro2018anchors, shrikumar2017learning} and visual heat maps~\cite{chattopadhay2018grad, gradCAM, wang2020score, CAM}. Such methods are critical in applications where understanding the influence of specific inputs is paramount.

Inherently interpretable models~\cite{rudin2019stop} are designed to provide direct insights into the causality within model predictions. These models facilitate a straightforward tracing of cause and effect in decision-making processes~\cite{bohle2022b, 
izza2020explaining, nauta2021neural, ribeiro2016should, sammani2022nlx}. Recent advancements have leveraged VLMs~\cite{alayrac2022flamingo,li2023blip,lu2019vilbert, CLIP} such as CLIP to enhance interpretability, exploiting the model's ability to measure the similarity between images and descriptive features directly~\cite{cui2023ceir, feng2023text, lewis2023gist, menon2022visual, yan2023learning, yang2023language}.

Extending interpretability in image classification tasks, the use of language descriptions to explain model predictions has been applied across various tasks including object recognition~\cite{li2024desco, nauta2021neural, park2018multimodal}, visual question answering~\cite{park2018multimodal, sammani2022nlx}, text classification~\cite{tan2024interpreting}, and image generation~\cite{hu2023tifa}. This approach often involves prompting LLMs~\cite{gpt-4,GPT-3,thoppilan2022lamda, touvron2023llama} to extract and generate relevant content and candidates for explanations~\cite{changpinyo2022all, cui2023ceir, feng2023text, hu2023tifa, lewis2023gist, li2024desco, menon2022visual,    yan2023learning, yang2023language}, bridging the gap between the need for extensive data annotations and human-understandable interpretations.

CBMs~\cite{koh2020concept} represent a structured approach to interpretability, where models explicitly predict a set of intermediary concepts before arriving at a final decision. Many studies have embraced this model structure~\cite{chauhan2023interactive, cui2023ceir, feng2023text, 
huang2024concept, 
sawada2022concept, semenov2024sparse, 
xu2024energy, 
yan2023learning, yang2023language}. 
To enhance the trustworthiness of the concepts, \citet{kim2023concept} introduces a visual activation score to assess whether a concept is visually represented. Another approach by \citet{lockhart2022learn} allows CBMs to abstain from predictions about concepts when there is insufficient confidence, thus focusing on enhancing model reliability when explanations lack certainty.

Some studies have applied a probabilistic approach for constructing CBMs, focusing on modeling the space of concepts within these models. One such study by \citet{kazmierczak2023clip-qda} represents the distribution of CLIP similarity scores by a mixture of Gaussians and employs statistical tools to explain the classifier. They integrate post-hoc methods such as LIME~\cite{ribeiro2016should} and SHAP~\cite{SHAP} to ascertain the importance scores of each concept. 
Another research by \citet{kim2023probabilistic} develops a CBM that predicts concept embeddings by sampling from a normal distribution with learned parameters. 
Merely predicting embeddings, however, can be inadequate for robust interpretation, as predictions often lack the necessary textual descriptions that anchor these embeddings.

The accurate modeling of the bottleneck involves grasping the distribution density of concepts. 
Estimating the PDF is a complex yet vital task, tackled by various probabilistic methods~\cite{goodfellow2014generative, kingma2013auto, power2002understanding,
rezende2015variational,   wkeglarczyk2018kernel}. Techniques such as flow networks utilize sequences of invertible transformations to refine a simple initial density into a more complex, desired distribution~\cite{rezende2015variational}. In score matching~\cite{score-matching}, by minimizing the discrepancy between the empirical and model-based score functions, one can learn the density function indirectly. 
This idea has gained attention across various applications and underpins the theoretical framework of diffusion-based generative models~\cite{ho2020denoising, sohl2015deep, song2019generative}.

%% file: Sec/3_method.tex
\section{Method}
\label{sec:method}

\subsection{Problem formulation}
\label{problem-formulation}

We outline the problem we aim to address and detail the approach in developing CLEAR. We consider a dataset $D$ consisting of images and their corresponding labels, denoted as $D=\{(i,c)\}$, where each class $c\in C$ is associated with a set of attributes $A_{c}=\{a_{c_{1}}, \dots , a_{c_{l}}\}$. We define the union of all these attribute sets as our descriptor pool $A=\bigcup_{c\in C}A_{c}$.
Using a VLM, equipped with text and image encoders $E_T$ and $E_I$, respectively, we project our dataset into the VLM's embedding space $\mathbb{R}^d$. Through these encoders, we denote the obtained embeddings for our images and for our descriptor pool
$I=\{E_I(i)|(i,c)\in D\} \text{ and } P=\{ E_T(a)|a\in A\}$.

Our method explores CBMs, which are models of the form $f(g(x))$ and comprise two functions: $g:\mathbb{R}^d \rightarrow \mathbb{R}^k$, mapping an input x to a concept space (the model's bottleneck), and $f:\mathbb{R}^k \rightarrow \mathbb{R}$, mapping data from the concept space to make the final prediction. 
The process of learning a CBM can follow one of three approaches: 
an independent bottleneck approach, where the functions $f$ and $g$ are learned separately with their respective loss functions minimized independently; 
a sequential bottleneck strategy, which involves training $f$ and $g$ separately by learning $g$ first, fixing it and then learning $f$; 
or a joint bottleneck technique, where $f$ and $g$ are trained simultaneously with the objective to minimize the combined sum of their losses. 

Our research adopts a sequential-bottleneck approach to CBM construction. Initially, we learn an approximation to the function $g$ (see Section \ref{embedding-approximation-learning}). Following this, we move on to formally construct the function $g$, in Section \ref{concept-selection}. Finally, once $g$ is established, we fix it and train a function $f$, detailed in Section \ref{bottleneck-integration}. This methodology, illustrated in Figure~\ref{fig:architecture}, ensures that each component of the model is optimally trained for its specific role in the overall architecture. 

\input{figures/architecture}
\subsection{Embedding approximation learning}
\label{embedding-approximation-learning}

Our initial goal is to construct an approximation of the conceptual bottleneck. We focus on learning a linear function, $S:\mathbb{R}^d\rightarrow \mathbb{R}^k$ , characterized by its weight matrix, $[S]\in \mathbb{R}^{d\times k}$.
This matrix serves as a collection of $k$  unique $d$-dimensional embeddings. 
Since $S$ transforms an image embedding into the concept space, training $S$ appropriately results in optimal embeddings that accurately embody our concepts.
To facilitate the learning of $S$, we train it along a function $W:\mathbb{R}^k\rightarrow \mathbb{R}^{|C|}$, responsible for making the model's classification.
Although these functions structurally resemble a CBM, this model is not yet a CBM as it lacks integration with the prior conceptual knowledge we aim to incorporate.

To facilitate the learning of meaningful embeddings, the joint image–descriptor distribution $p(x)$ is modeled into another distribution $p_\theta(x)$ using score matching. 
The scores of the joint image–descriptor distribution $s_\theta(x)$ are learned such that $s_\theta(x) = \nabla_x \log p_\theta(x)\approx \nabla_x \log p(x)$, aiming to minimize the Fisher Divergence $D(p||p_\theta) = \frac{1}{2}\mathbb{E}_{p(x)}\left[||\nabla_x\log p(x) - s_\theta (x)||_2^2\right]$. 
As demonstrated by \citet{score-matching}, this is equivalent to minimizing $\mathbb{E}_{p(x)}\left[\frac{1}{2} ||s_\theta (x) ||_2^2 +
tr(\nabla_x s_\theta (x))\right]$.
Due to the computational complexity of the Hessian term $\nabla_x s_\theta (x)$, sliced score matching (SSM), where scores are projected onto a random direction $v\sim p_v$, is employed.  
As shown by \citet{slicedSM}, minimizing the following provides a computationally viable alternative to minimizing the Fisher Divergence.

\begin{equation}
\label{eq:ssm-objective}
\mathbb{E}_p\mathbb{E}_{p_v}\left[v^T\nabla_x s_\theta (x) v + 
\frac{1}{2} ||s_\theta (x)||_2^2\right]
\end{equation}

We obtain the score function and integrate it into the embedding learning process using Langevin dynamics. This approach steers the concept embeddings toward the high-density areas of $p(x)$, thereby better aligning with the data’s images and descriptions. 
The learning of the embeddings in $[S]$ is constrained using the $L_2$ norm toward the transformed embeddings.
The Langevin dynamics' sampling procedure iteratively updates an example $x_0$ by $x_{i+1}\gets x_i + \epsilon \nabla_x \log p(x_i) + \sqrt{2\epsilon}z_i$ where $z_i\sim \mathcal{N}(0,1)$ for $i=0,\dots , t$.
This procedure is applied to the columns of $[S]$, transforming a learned embedding $[S]^T_j$, as follows: 

\begin{equation}
\label{eq:sampling}
{[S]^T_j}_{(i+1)}\gets {[S]^T_j}_{(i)} + \frac{\epsilon}{2} \cdot s_\theta ({[S]^T_j}_{(i)}) + \sqrt{\epsilon}z_i
\end{equation}

The overall term guiding the learning of $[S]$ toward $p(x)$ is:

\begin{equation}
\label{eq:sm-loss}
\mathcal{L}_{SM}(S) = \frac{1}{k}\sum^{k}_{j=1} ||{[S]^T_j}_{(t)} - [S]^T_j||_2^2
\end{equation}

To ensure accurate classification, the cross-entropy loss function is employed, defined as
$\mathcal{L}_{CE}(x,c)=\sum^{|C|}_{j=1}\delta _{j,c}\cdot \log \left(W(S(x))_j\right)$,
where  $W(S(x))_j$ denotes the model's predicted probability for class $j$ and $\delta_{j,c}$ 
is the indicator of whether class $c$ is the correct classification for image $x$.
The composite loss function that guides the optimization of our model is formulated as follows:

\begin{equation}
\label{eq:loss-function}
\mathcal{L}(S,x,c)=\lambda \cdot \mathcal{L}_{SM}(S) + \mathcal{L}_{CE}(x,c) = \frac{\lambda}{k}\sum\limits^{k}_{j=1} ||{[S]^T_j}_{(t)} - [S]^T_j||_2^2 + \sum\limits^{|C|}_{j=1}\delta _{j,c}\cdot \log \left(W(S(x))_j\right)
\end{equation}

Thus, for a given image embedding $x$, our model determines its prediction by:
\[\hat{c}=argmax(W(S(x)))\]

\subsection{Concept selection}
\label{concept-selection}

The essence of interpretability in a CBM is rooted in textual concepts, 
meaning merely deriving the bottleneck approximation $S$ is insufficient.
Thus, we venture further, identifying from $P$ a subset of descriptors that closely align with $S$. 
Using a matrix $M_p\in \mathbb{R}^{|P|\times d}$ whose rows are the elements of $P$, we construct an approximation-description similarity matrix $Sim \in \mathbb{R}^{k\times |P|}$ by:

\begin{equation}
\label{eq:sim}
Sim=[S]^T \cdot M_p^T
\end{equation}

Here, $Sim_{i,j}$ measures the similarity between the \textit{i-th} approximation in $[S]$ and the \textit{j-th} descriptor in $P$.
Our aim is to uniquely pair each of the $k$ conceptual approximations with a descriptor, maximizing their joint similarity. To achieve this, we employ the Hungarian method~\cite{kuhn1955hungarian}, an algorithm that finds the optimal assignment that minimizes a total cost in a bipartite matching scenario. Inverting this, we turn the goal of maximizing similarity into one of minimizing cost by subtracting each matrix entry from its highest value, thereby aiming to minimize the total deviation from the maximum similarity. 

The runtime complexity of the Hungarian algorithm on $Sim$ is $O(|P|^3)$, which becomes computationally challenging when the descriptor pool is large. To enhance the efficiency, we refine the process by retaining only the top $m$-most similar concepts in the similarity matrix for each learnable concept. 
This approach reduces the pool size significantly while still preserving the most promising candidates for matching.
Applying this approach enables us to identify an optimal set of embeddings and generate a new matrix, $[S]'\in \mathbb{R}^{d\times k}$, which includes the selected descriptors from $P$. 

\subsection{Bottleneck integration}
\label{bottleneck-integration}

Building on the foundation laid in Sections~\ref{problem-formulation} to~\ref{concept-selection}, where we derived the matrix $[S]'$, we now establish the function $S':\mathbb{R}^d \rightarrow \mathbb{R}^k$. 
This function transforms an image embedding $x$ into the concept space through $S'(x)=[S]'\cdot x$. This function $S'$ is exactly the function $g:\mathbb{R}^d \rightarrow \mathbb{R}^k$ of our CBM.
To determine the label prediction function $f:\mathbb{R}^k \rightarrow \mathbb{R}$, we fix $g$ and proceed to learn a linear function $W':\mathbb{R}^k \rightarrow \mathbb{R}^{|C|}$, in line with the approach in Section~\ref{embedding-approximation-learning}.
In this instance, however, we streamline the learning process by using the cross-entropy loss $CE(x,c)$ alone, focusing on guiding the transformation of data from the concept space to a classification.
The prediction function $f$ is then defined as $f(x')=argmax(W'(x'))$. By integrating $g$ and $f$, we synthesize the complete CBM:
\[f(g(x))=argmax(W'(S'(x))\]

%% file: figures/architecture.tex
\begin{figure*}[t]
    \centering
    \includegraphics[width=\textwidth]{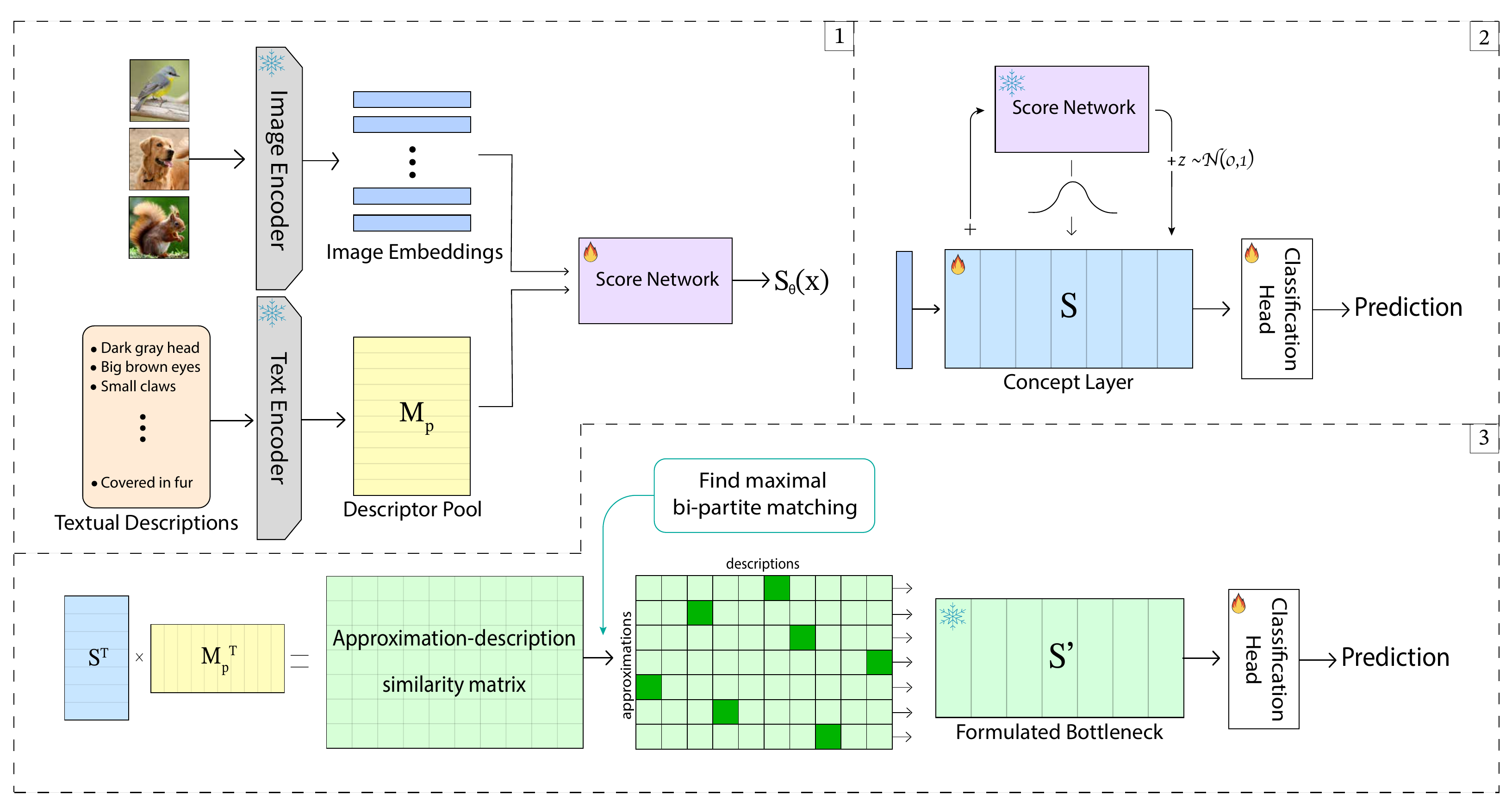}
    \caption{An overview of CLEAR. In step 1 we obtain the image and descriptor embeddings and train the score network. 
    In step 2 we learn the concept approximations 
    and in step 3 we obtain the approximation-description similarity matrix, select the concepts by finding the optimal allocation and integrate our bottleneck.}
    \label{fig:architecture}
\end{figure*}

%% file: Sec/4_experiments.tex
\section{Experiments}
\label{sec:experiments}

\input{tables/all-results}

\subsection{Experimantal setup}
\label{experimental-setup}

\textbf{Implementation details}
In our research, we used the CLIP~\cite{CLIP} package developed by OpenAI, specifically adopting the base architecture of ViT-B/32 for image encoding. The descriptor pool employed is the textual descriptions provided in LaBo~\cite{yang2023language}, generated by GPT-3~\cite{GPT-3} and filtered per class.
The score function was constructed using a network of three linear layers.
Model training leveraged the PyTorch library and the Adam~\cite{kingma2014adam} optimizer for efficient optimization.

For learning the embedding approximations detailed in Section~\ref{embedding-approximation-learning}, we selected model checkpoints that achieved the highest validation accuracy for use in the concept selection phase outlined in Section~\ref{concept-selection}.
Similarly, we chose the model with the highest validation accuracy, as seen in Section~\ref{bottleneck-integration}, and reported its results.
All experiments were conducted using an Nvidia GeForce RTX 3090, and the hyperparameters for all datasets and configurations are provided in the supplementary material in Section~\ref{appendix/implementation-details}.

\textbf{Baselines} We compared our method with two other interpretable CBM strategies—LaBo~\cite{yang2023language} and the framework developed by \citet{yan2023learning}, elaborated upon in Section~\ref{sec:intro}. 
LaBo~\cite{yang2023language} requires that the bottleneck size be a multiple of the dataset's class count. 
Therefore, for each dataset containing $|C|$ classes, we assessed our model using bottleneck sizes of $|C|$ and $2|C|$.
Additionally, we included an experiment with a smaller bottleneck in line with \citet{yan2023learning},
validating our findings and demonstrating the efficacy of identifying a precise set of concepts.
This approach ensures a thorough evaluation and showcases conceptual clarity and efficiency improvement.

\textbf{Datasets} 
Our framework for image classification was evaluated across five datasets: CIFAR-10~\cite{CIFAR10-and-CIFAR100},
CIFAR-100~\cite{CIFAR10-and-CIFAR100},
CUB-200-2011~\cite{WahCUB_200_2011},
Flower-102~\cite{Flower} 
and Food-101~\cite{bossard2014food}.
Training was conducted on the full datasets, with performance assessed based on the accuracy obtained from the test sets.

\subsection{Main results}
\label{main-results}
The results of the aforementioned experiments
are presented in Table~\ref{all-results}.
We conducted three independent runs for each experimental condition to ensure the robustness of our findings. 
We report the mean accuracy and the standard error of the mean of these runs, providing a view of the variability and reliability of our method.
Across the benchmarks, our method demonstrates an average performance increase of 2.84\%, with specific gains in accuracy noted across different datasets: a 2.99\% increase in CIFAR-10, 0.47\% in CIFAR-100, 3.75\% in Flower, 5.78\% in CUB, and 1.23\% in Food. 
These improvements underscore our method's effectiveness and highlight our approach's adaptability. The full results of the independent runs are listed in Section~\ref{appendix/additional-results} in the appendix.

\paragraph{Bottleneck size}

To explore the influence of the number of concepts selected, we conducted experiments with varying concept quantities: 8, 16, 32, 64, 128, 256, and the entire descriptor pool (findings are presented in Figure~\ref{fig:bottleneck-sizes-graphs}).
The findings generally indicate that using more concepts tends to enhance data representation. 
More specifically, in CIFAR-10, our model with 256 concepts outperforms the model using the full set of 500 descriptors. 
Similarly, in CUB, using 64 concepts yields better results than the full pool of 10,000 descriptors, and in the Food dataset, selecting 202 concepts surpassed the performance of the full pool containing 5,050 descriptors. These observations suggest that our method effectively identifies the most relevant concepts for each dataset and excludes less pertinent descriptors. 
Considering the CUB dataset as an example, the superior performance observed at intermediate concept sizes can potentially be explained by the homogeneity of the classes, with all focusing on various bird species. This conceptual similarity among the classes means fewer annotations are necessary for adequate interpretation.

\input{figures/bottleneck-sizes}

\paragraph{Interpretability analysis}

To assess the interpretability of our model during its construction and at inference time, we examined how it forms and uses its concepts at different bottlenecks across four datasets: CIFAR-10, CIFAR-100, Flower, and Food. 
For each dataset, we identified the bottleneck formed and reported some of the key concepts presented in it. 
At inference, we present the normalized concept scores for a selected image from each dataset and pinpoint the most significant concept, offering a clear and effective way to interpret the model’s decisions.
For instance, an image of a truck from CIFAR-10 shows the highest similarity with the concept \textit{“large, red vehicle with four wheels”}, and an image of a Caesar salad from Food shows the highest similarity with \textit{“green salad with romaine lettuce, croutons, and Parmesan cheese”}.
The selected concepts presented in Figure~\ref{fig:inference-fig} offer informative interpretations. 
These examples highlight how our model's interpretability not only aids in understanding its internal mechanisms but also enhances trust in its outputs by aligning its decision pathways with comprehensible and relevant human-understandable concepts.
Additional visualizations reflecting the diversity of the bottleneck are presented in Section~\ref{appendix/visualizations} in the appendix.

\subsection{Ablation study}

To dissect the contribution of individual components within our model, an ablation study was conducted, focusing on four key aspects:

\paragraph{Regularization}
In our approach, the regularization function steers the learning process toward the joint image–descriptor distribution, calculated by Equation~\ref{eq:sm-loss} and weighted into the complete loss function in Equation~\ref{eq:loss-function}.
We evaluate its significance by contrasting it with alternative methods. One such method involves the Mahalanobis distance~\cite{mahalanobis2018generalized}, a statistical measure that evaluates the distance from a point to a distribution.
The Mahalanobis distance between a point $x\in \mathbb{R}^n$ and a distribution $Q$ with mean and covariance matrix $\mu \in \mathbb{R}^n$ and $\Sigma\in \mathbb{R}^{n\times n}$ is $\sqrt{(x-\mu)^T\cdot \Sigma ^{-1} \cdot (x-\mu)}$.
In our context, this distance between each row in $[S]^T$ and the embedding space distribution approximated by $P$ is defined as $\mathcal{L}_{MA}(S)=\frac{1}{k} \sum^{k}_{j=1} \sqrt{([S]^T_j-\mu)^T\cdot \Sigma ^{-1} \cdot ([S]^T_j-\mu)}$, where $\mu$ and $\Sigma$ are the mean and covariance of $P$.
Additionally, we compare this to using the Euclidean distance ($L_2$ norm), which computes the distance between each learned embedding and the descriptor embeddings as 
$\mathcal{L}_{EU}(S) = \sum^{k}_{j=1}\frac{1}{|P|}\sum^{|P|}_{h=1}||[S]^T_j-e_h||_2^2$ for every $e_h \in P$.
Furthermore, we assess the impact of omitting any regularization, focusing learning solely on classification via cross-entropy.

We evaluate the effectiveness of each regularization approach by conducting a comparison among these methods. 
The results, as outlined in Table~\ref{regularization-ablation-table}, demonstrate the superiority of our proposed learning method over the alternatives.

\input{figures/inference}

\input{tables/regularization-ablation}

\paragraph{Pool size} We evaluated the impact of the size of the descriptor pool by comparing the pool used in our method with the one used in~\citet{yan2023learning}. 
The findings in Table~\ref{pool-size-ablation-table} support the idea that having a more extensive pool of descriptors enhances our ability to match the descriptor distribution closely.

\input{tables/pool-size-ablation}

\paragraph{Concept selection}
We assessed the effectiveness of our concept selection strategy by evaluating the Hungarian algorithm~\cite{kuhn1955hungarian} that we implemented in our approach, as outlined in Section~\ref{concept-selection}.
This method is initially compared to the nearest neighbor (NN) algorithm, similar to the approach taken by \citet{yan2023learning}, which matches each learned embedding with the nearest concept based on the $L_2$ distance.
We also contrasted this with a strategy involving randomly selecting descriptors from our available pool. 
This comparative analysis ensures that our method's performance enhancements are meaningful and substantiated. It also justifies the complexity added by the Hungarian algorithm over simpler or more random approaches.
The figures in Table~\ref{selection-ablation-table} show that choosing concepts by maximizing the embeddings' joint similarity yields better results compared to other approaches.

\input{tables/concept-selection-ablation}

%% file: tables/all-results.tex
\begin{table}[t!]
  \caption{The comparison of the proposed model on various benchmarks compared to baseline state-of-the-art methods.}
  \label{all-results}
  \centering

\resizebox{\textwidth}{!}{
\begin{tabular}{c cll cll cll}
\toprule
        & \multicolumn{9}{c}{Datasets} \\
       \cmidrule(r){2-10}
          & \multicolumn{3}{c}{CIFAR-10}                                                  & \multicolumn{3}{c}{CIFAR-100}                                                 & \multicolumn{3}{c}{Flower}                 \\ 
  Bottleneck Size  & 8                         & \multicolumn{1}{c}{10}   & \multicolumn{1}{c}{20} & 64                        & \multicolumn{1}{c}{100} & \multicolumn{1}{c}{200} & 32                        & \multicolumn{1}{c}{102} & \multicolumn{1}{c}{204} \\ 
         \hline\noalign{\smallskip}\hline\noalign{\smallskip}

LaBo~\cite{yang2023language}            & -                         & \multicolumn{1}{c}{78.11}                     & \multicolumn{1}{c}{84.84}                   & -                         & \multicolumn{1}{c}{75.10}                    & \multicolumn{1}{c}{76.94}                     & -                         & \multicolumn{1}{c}{80.98}                    & \multicolumn{1}{c}{86.76}                       \\
Yan et al.~\cite{yan2023learning}      & 77.47 & \multicolumn{1}{c}{80.09 }                   & \multicolumn{1}{c}{87.99}                   & 73.31 & \multicolumn{1}{c}{75.12}                   & \multicolumn{1}{c}{77.29}          & 80.88 & \multicolumn{1}{c}{87.26}                   & \multicolumn{1}{c}{89.02}                    \\
\rowcolor{golden!10} CLEAR      & \textbf{81.17$\pm$0.58}                      & \multicolumn{1}{c}{\textbf{84.19$\pm$1.17}} & \textbf{89.16$\pm$0.73}                   & \multicolumn{1}{l}{\textbf{73.75$\pm$0.09}} & \textbf{76.07$\pm$0.03}                   & \textbf{77.32$\pm$0.00}                    & \multicolumn{1}{l}{\textbf{87.11$\pm$0.28}} & \textbf{90.19$\pm$0.19}                   & \textbf{91.10$\pm$0.05} \\
\bottomrule
                  
\end{tabular}

}
\end{table}

\begin{table}[t!]
  \centering
\resizebox{0.75\textwidth}{!}{
\begin{tabular}{c cll cll}
\toprule
        
 & \multicolumn{6}{c}{Datasets} \\
       \cmidrule(r){2-7}
          & \multicolumn{3}{c}{CUB} & \multicolumn{3}{c}{Food} \\ 
  Bottleneck Size & 32                        & \multicolumn{1}{c}{200} & \multicolumn{1}{c}{400} & 64                                 & \multicolumn{1}{c}{101} & \multicolumn{1}{c}{202} \\ 
         \hline\noalign{\smallskip}\hline\noalign{\smallskip}

LaBo~\cite{yang2023language}            & -                         & \multicolumn{1}{c}{60.93}                   & \multicolumn{1}{c}{62.61}                    & -                                  & \multicolumn{1}{c}{79.95}                   & \multicolumn{1}{c}{81.33}                    \\
Yan et al.~\cite{yan2023learning}      &  60.27 & \multicolumn{1}{c}{63.88}                  & \multicolumn{1}{c}{64.05}                    & 78.41 & \multicolumn{1}{c}{80.22}                   & \multicolumn{1}{c}{81.85}           \\
\rowcolor{golden!10} CLEAR      & \textbf{65.42$\pm$0.17}                      & \textbf{70.18$\pm$0.14}         & \multicolumn{1}{c} {\textbf{69.94$\pm$0.13}}  & \multicolumn{1}{l}{\textbf{79.79$\pm$0.07}}         & \textbf{81.61$\pm$0.15}         & \textbf{82.77$\pm$0.18} \\
\bottomrule
\end{tabular}
}
\end{table}

%% file: figures/bottleneck-sizes.tex
\begin{figure*}[t]
    \centering
    \includegraphics[width=\textwidth]{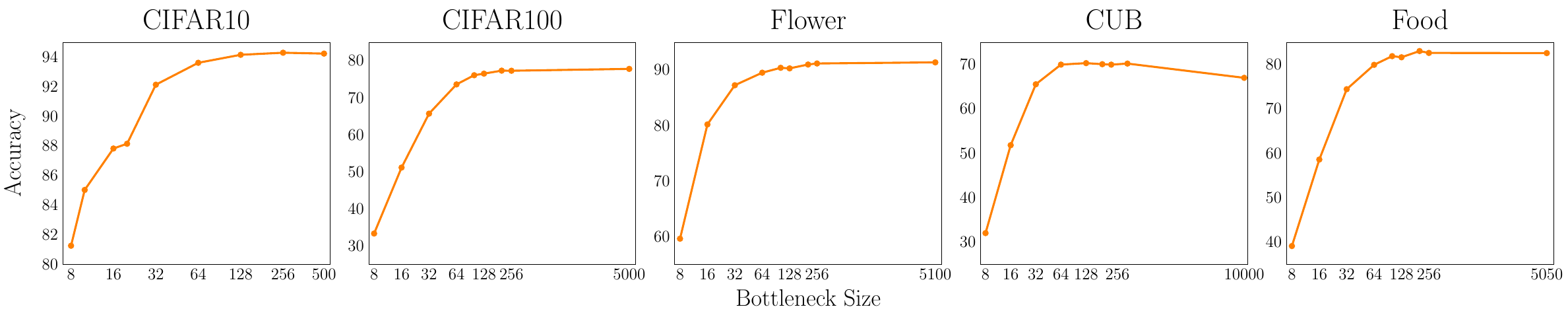}
    \caption{Test accuracy comparison on different bottleneck sizes across all datasets.}
    \label{fig:bottleneck-sizes-graphs}
\end{figure*}

%% file: figures/inference.tex
\begin{figure*}[t]
    \centering
    \includegraphics[width=\textwidth]{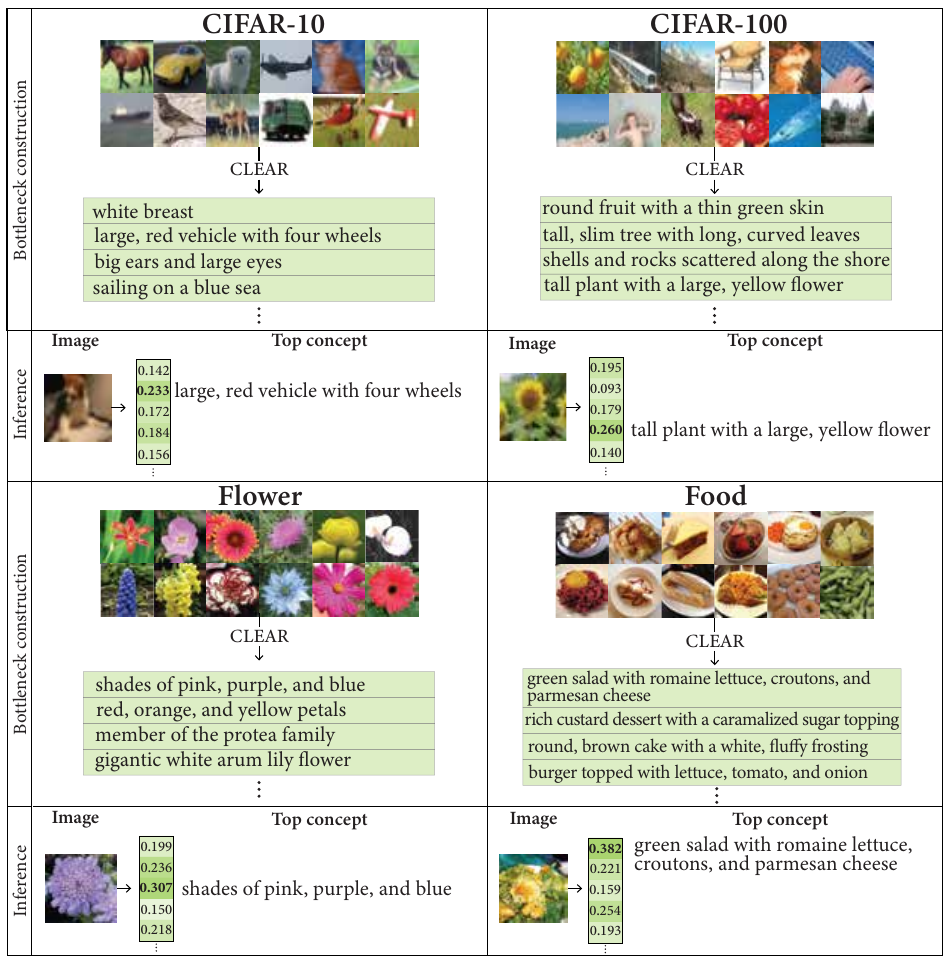}
    \caption{Interpretability analysis of our proposed framework during inference}
    \label{fig:inference-fig}
\end{figure*}

%% file: tables/regularization-ablation.tex
\begin{table}[h]
\caption{An examination of the impact of the regularization function.}
\label{regularization-ablation-table}
\centering
\begin{tabular}{ccccccccc}
\toprule
 & \multicolumn{8}{c}{No. of concepts in Food} \\
 \cmidrule(r){2-9}
 Regularization  & 8   & 16  & 32  & 64  & 128 & 256 & 101 & 202\\ \hline \noalign{\smallskip}\hline\noalign{\smallskip}

None             & 35.09 & 56.06 & 71.41 & 78.03 & 80.55 & 81.38 & 80.42 & 81.00 \\
$\mathcal{L}_{EU}$           & 23.41 & 44.07 & 64.45 & 76.50 & 80.97 & 82.51 & 80.18 & 82.35 \\
  $\mathcal{L}_{MA}$            & 33.46 & 55.13 & 71.18 & 77.93 & 80.48 & 81.32 & 80.12 & 81.35 \\
   \rowcolor{golden!10} $\mathcal{L}_{SM}$          & \textbf{39.10} & \textbf{58.58} & \textbf{74.42} & \textbf{79.91} & \textbf{81.61} & \textbf{82.59} & \textbf{81.86} & \textbf{83.01} \\

\bottomrule
\end{tabular}
\end{table}

%% file: tables/pool-size-ablation.tex
\begin{table}[h]
\caption{A study on the impact of the size of the descriptor pool}
\label{pool-size-ablation-table}
\centering
\begin{tabular}{cccccccccc}
\toprule
 & \multicolumn{9}{c}{No. of concepts in Flower} \\
 \cmidrule(r){2-10}
 Pool size & 8   & 16  & 32  & 64  & 128 & 256 & 102 & 204 & full \\ \hline \noalign{\smallskip}\hline\noalign{\smallskip}

$503$             & 54.41 & 75.88 & 80.09 & 85.29 & 88.72 & 89.90 & 87.84 & 90.09 & 91.17 \\
\rowcolor{golden!10} $5100$            & \textbf{59.60} & \textbf{80.19} & \textbf{87.25} & \textbf{89.51} & \textbf{90.29} & \textbf{91.17} &
\textbf{90.39} &
\textbf{90.98} &
\textbf{91.37} \\
\bottomrule
\end{tabular}
\end{table}

%% file: tables/concept-selection-ablation.tex
\begin{table}[h]
\caption{An analysis of the effects of the concept selection method}
\label{selection-ablation-table}
\centering
\begin{tabular}{ccccccccc}
\toprule
 & \multicolumn{8}{c}{No. of concepts in Flower} \\
 \cmidrule(r){2-9}
 Selection method  & 8   & 16  & 32  & 64  & 128 & 256 & 102 & 204 \\ \hline \noalign{\smallskip}\hline\noalign{\smallskip}

Random            & 20.19 & 51.96 & 70.19 & 79.90 & 86.76 & 88.72 & 85.39 & 88.52
\\
NN           & \textbf{59.60} & \textbf{80.19} & 87.15 & 88.92 & 89.70 & 90.88 & 89.70 & 90.68
\\
\rowcolor{golden!10} Our method             & \textbf{59.60} & \textbf{80.19 }& \textbf{87.25} &\textbf{ 89.51} & \textbf{90.29} & \textbf{91.17} & \textbf{90.39} & \textbf{90.98} \\

 \bottomrule
\end{tabular}
\end{table}

%% file: Sec/5_limitation.tex
\section{Limitation}
\label{sec:limitation}
The utilization of a sufficiently large descriptor pool is important for achieving accurate approximations, as well as for ensuring diversity of descriptions and adaptability to the data. 
It is, nevertheless, important to recognize that relying on a predefined descriptor pool can limit the applicability of the concept selection process to the specific dataset being used. 
While our work provides an optimal concept assignment, future research could enhance this field by developing methods to generate new concepts, further improving the descriptive power of datasets.
In addition, it should be considered that using LLM-generated concepts may introduce biases, which can reflect and perpetuate existing societal biases. This underscores the importance of being mindful of the broader societal impacts when leveraging LLMs for concept generation.

%% file: Sec/6_conclusion.tex
\section{Conclusion}
\label{sec:conclusion}
In this study, we developed a data-adaptive CBM by learning a precise set of concepts that accurately represents the data, thereby enabling clear interpretations and facilitating human interaction with the model. 
Our framework, CLEAR, was compared across five different image classification datasets, consistently achieving state-of-the-art results.
To thoroughly assess the contributions of our paradigm, we conducted an in-depth analysis of our model's components. 
We demonstrated the advantages of the individual components, each of which plays a crucial role in enhancing the performance and interpretability of our CBM.

While our method is evaluated on image classification tasks, we note that our approach directly applies to any computer vision task. This broad applicability stems from the versatility of VLMs when handling a wide range of tasks within this domain. Furthermore, adapting our framework with modifications suitable for different data types can broaden its application to various other tasks. For example, incorporating pre-trained language models as text encoders allows our framework to seamlessly adapt to text classification tasks.

%% file: Sec/8_appendix.tex
\newpage

\appendix

\section{Experiments}

\subsection{Additional results}
\label{appendix/additional-results}
We present comprehensive results that expand on those reported in the paper. Initially, we list the hyperparameter values used in our model configuration and present the test accuracy results for varying numbers of concepts: 8, 16, 32, 64, 128, 256, one per class (1-\textit{pc}), two per class (2-\textit{pc}), and the entire descriptor pool. 
We detail the hyperparameters used and their selection in Section~\ref{appendix/implementation-details} and report their results in Table~\ref{hyperparameters-table}.

Additionally, Table~\ref{full-all-results} includes the complete results of the three runs that form the basis for the mean and standard error reported for each dataset in Table~\ref{all-results}.

\bigskip

\input{tables/full-all-results}

\subsection{Implementation details}
\label{appendix/implementation-details}

\paragraph{Hyperparameters}
In Table~\ref{hyperparameters-table}, we provide an overview of the hyperparameters that configure our model for each dataset, along with their values and the empirical results. During the sampling procedure from the joint image–descriptor distribution, the transformation is calculated as formulated in Equation~\ref{eq:sampling}.
For each dataset, we search for suitable values of $\epsilon \in \{1, 0.1, 0.01\}$ and $t \in \{1, 3, 5, 7, 10\}$. To balance the two terms in the loss function, we determine the optimal $\lambda$ value from $\lambda \in \{1, 0.1, 0.01, 0.001\}$.

We also fine-tune the batch size, learning rate, random seed values, and the number of epochs for training the model during the embedding approximation learning phase ($\textit{epochs}_1$), as detailed in Section~\ref{embedding-approximation-learning}, and during the training of the linear layer ($\textit{epochs}_2$), as described in Section~\ref{bottleneck-integration}.

We obtain our score model by minimizing the objective presented in Equation~\ref{eq:ssm-objective}. Training the score model involves a network with three linear layers, each having hidden dimensions of 1024. For all datasets, training is performed on the image and descriptor embeddings for 1000 epochs using the Adam optimizer with a fixed learning rate of $1e-4$. The batch size for the images remains the same as before, while the batch size for the descriptors is set to 32.

\input{tables/hyperparameters}

\paragraph{Descriptors pool filtering}
During the concept selection phase described in Section~\ref{concept-selection}, we contruct $Sim$ by employing Equation~\ref{eq:sim} and retain only the top $m$-most similar concepts in $Sim$ for each learnable concept. Initially, we set $m$ to 5 and define $TopDes = \bigcup_{i=1}^k \{sort(Sim_{i})^{(1)}, \dots , sort(Sim_{i})^{(m)} \}$, where we sort $Sim_{i}$ and select the top $m$-most similar embeddings. If the resulting pool size is greater than $k$, we proceed with concept selection; otherwise, we iteratively find $TopDes$ for $m_{i+1}=2m_i$ until this condition is met. Generally, a low value of $m$ indicates diverse learned embeddings. The reader is reffered to Table~\ref{hyperparameters-table} for the obtained values.

\paragraph{Citations and rights}
We have thoroughly cited all datasets and research papers used in our experiments throughout our paper. The CLIP model~\cite{CLIP} is available under the MIT license.

\section{Descriptor visualizations}
\label{appendix/visualizations}
To gain insights into the structure of the textual descriptions, we visualize the descriptor pool along with the selected concepts that form our bottleneck. This visualization allows us to understand the diversity in the selection of the concepts.

By lowering the dimension of each embedding, we use t-SNE~\cite{t-SNE} to visualize both the embeddings of the descriptor pool and the embeddings of the selected concepts. The visualizations for the CIFAR-10, CIFAR-100, Flower, CUB, and Food datasets are presented in Figures~\ref{fig:cifar10-atts-vis} to~\ref{fig:food-atts-vis}. In these visualizations, each green point represents a concept from the descriptor pool, while each blue point represents a concept in the CLEAR bottleneck. 

These visualizations illustrate how well the selected concepts represent the broader pool, which is vital for ensuring the robustness and generalizability of our approach. They demonstrate that our method's concept selections effectively distinguish between different conceptual areas and provide a diverse set of concepts.

\input{figures/visualizations}

%% file: tables/full-all-results.tex
\begin{table}[h]
\caption{Complete results of the three runs for each dataset}
\label{full-all-results}
\centering

\begin{tabular}{c*{3}{>{\centering\arraybackslash}p{2cm}}}
\toprule
 \multirow{2}{*}{Dataset} & \multicolumn{3}{c}{\multirow{2}{*}{\parbox{4cm}{\centering Accuracy when varying \\ no. of concepts}}} \\
 & & & \\
 \hline \noalign{\smallskip}\hline\noalign{\smallskip}

\multirow{4}{*}{CIFAR-10}  & \textbf{8} & \textbf{10} & \textbf{20} \\
  \cmidrule(lr){2-2} 
  \cmidrule(lr){3-3} 
  \cmidrule(lr){4-4}  
  & 81.25 & 85.02 & 88.14\\  
 & 80.11  & 85.68 & 88.73 \\ 
 & 82.16 & 81.87 & 90.61 \\ \hline \noalign{\smallskip}\hline\noalign{\smallskip}

 \multirow{4}{*}{CIFAR-100} & \textbf{64} & \textbf{100} & \textbf{200} \\
  \cmidrule(lr){2-2} 
  \cmidrule(lr){3-3} 
  \cmidrule(lr){4-4} 
   & 73.6 & 76.08 & 77.32\\  
 & 73.71  & 76.12 & 77.31 \\
 & 73.94 & 76.01 & 77.33 \\ \hline \noalign{\smallskip}\hline\noalign{\smallskip}

  \multirow{4}{*}{Flower}  & \textbf{32} & \textbf{102} & \textbf{204} \\
  \cmidrule(lr){2-2} 
  \cmidrule(lr){3-3} 
  \cmidrule(lr){4-4} 
  & 87.25 & 90.39 & 90.98 \\  
 & 86.56  & 90.39 & 91.17 \\
 & 87.54 & 89.80 & 91.17 \\ \hline \noalign{\smallskip}\hline\noalign{\smallskip}

   \multirow{4}{*}{CUB}  & \textbf{32} & \textbf{200} & \textbf{400} \\
  \cmidrule(lr){2-2} 
  \cmidrule(lr){3-3} 
  \cmidrule(lr){4-4} 
  & 65.53 & 70.05 & 70.19 \\  
 & 65.08  & 70.48 & 69.71 \\
 & 65.67 & 70.02 & 69.93 \\ \hline \noalign{\smallskip}\hline\noalign{\smallskip}

  \multirow{4}{*}{Food}  & \textbf{64} & \textbf{101} & \textbf{202} \\
  \cmidrule(lr){2-2} 
  \cmidrule(lr){3-3} 
  \cmidrule(lr){4-4} 
   & 79.91 & 81.86 & 83.01 \\  
 & 79.83  & 81.64 & 82.40 \\
 & 79.64 & 81.33 & 82.92 \\ 
 \bottomrule
\end{tabular}
\end{table}

%% file: tables/hyperparameters.tex
\begin{table}[!h]
\caption{Hyperparameter values and full results on varying numbers of concepts.}
\label{hyperparameters-table}
\centering
\resizebox{0.815\textwidth}{!}{
\begin{tabular}{c ccccccccc}
\toprule
 Dataset & \multicolumn{9}{c}{CIFAR-10} \\
 \hline \noalign{\smallskip}\hline\noalign{\smallskip}

 $\epsilon$ & \multicolumn{9}{c}{1} \\
 
  $t$ & \multicolumn{9}{c}{7} \\

$\lambda$ & \multicolumn{9}{c}{0.01} \\

 batch size & \multicolumn{9}{c}{4096} \\

learning rate & \multicolumn{9}{c}{0.01} \\

seed  & \multicolumn{9}{c}{4} \\

 $\textit{epochs}_1$ & \multicolumn{9}{c}{1000} \\

$\textit{epochs}_2$ & \multicolumn{9}{c}{2000} \\

 \hline

no. of concepts & 8 & 16 & 32 & 64 & 128 & 256 & 1-\textit{pc} & 2-\textit{pc} & full  \\

$m$ & 5 & 5 & 5 & 5 & 5 & 10 & 5 & 5 & - \\

accuracy & 81.25 & 87.82 & 92.13 & 93.61 & 94.15 & 94.29 & 85.02 & 88.14 & 94.23 \\

\toprule
 Dataset & \multicolumn{9}{c}{CIFAR-100} \\
 \hline \noalign{\smallskip}\hline\noalign{\smallskip}

 $\epsilon$ & \multicolumn{9}{c}{0.1} \\
 
  $t$ & \multicolumn{9}{c}{5} \\

$\lambda$ & \multicolumn{9}{c}{0.1} \\

 batch size & \multicolumn{9}{c}{4096} \\

learning rate & \multicolumn{9}{c}{0.01} \\

seed  & \multicolumn{9}{c}{0} \\

 $\textit{epochs}_1$ & \multicolumn{9}{c}{1000} \\

$\textit{epochs}_2$ & \multicolumn{9}{c}{4000} \\

 \hline

no. of concepts & 8 & 16 & 32 & 64 & 128 & 256 & 1-\textit{pc} & 2-\textit{pc} & full  \\

$m$ & 5 & 5 & 5 & 5 & 5 & 5 & 5 & 5 & - \\

accuracy & 33.30 & 51.13 & 65.7 & 73.6 & 76.51 & 77.29 & 76.08 & 77.32 & 77.79 \\

\toprule
 Dataset & \multicolumn{9}{c}{Flower} \\
 \hline \noalign{\smallskip}\hline\noalign{\smallskip}

 $\epsilon$ & \multicolumn{9}{c}{0.1} \\
 
  $t$ & \multicolumn{9}{c}{5} \\

$\lambda$ & \multicolumn{9}{c}{0.01} \\

 batch size & \multicolumn{9}{c}{4096} \\

learning rate & \multicolumn{9}{c}{0.001} \\

seed  & \multicolumn{9}{c}{1} \\

 $\textit{epochs}_1$ & \multicolumn{9}{c}{2000} \\

$\textit{epochs}_2$ & \multicolumn{9}{c}{20000} \\

 \hline

no. of concepts & 8 & 16 & 32 & 64 & 128 & 256 & 1-\textit{pc} & 2-\textit{pc} & full  \\

$m$ & 5 & 5 & 5 & 5 & 5 & 5 & 5 & 5 & - \\

accuracy & 59.60 & 80.19 & 87.25 & 89.51 & 90.29 & 91.17 & 90.39 & 90.98 & 91.37 \\

\toprule

 Dataset & \multicolumn{9}{c}{CUB} \\
 \hline \noalign{\smallskip}\hline\noalign{\smallskip}

 $\epsilon$ & \multicolumn{9}{c}{1} \\
 
  $t$ & \multicolumn{9}{c}{10} \\

$\lambda$ & \multicolumn{9}{c}{1} \\

 batch size & \multicolumn{9}{c}{32} \\

learning rate & \multicolumn{9}{c}{0.01} \\

seed  & \multicolumn{9}{c}{0} \\

 $\textit{epochs}_1$ & \multicolumn{9}{c}{5000} \\

$\textit{epochs}_2$ & \multicolumn{9}{c}{8000} \\

 \hline

no. of concepts & 8 & 16 & 32 & 64 & 128 & 256 & 1-\textit{pc} & 2-\textit{pc} & full  \\

$m$ & 5 & 5 & 5 & 5 & 5 & 5 & 5 & 5 & - \\

accuracy & 32.01 & 51.81 & 65.53 & 69.96 & 70.29 & 69.95 & 70.05 & 70.19 & 66.98 \\

\toprule

 Dataset & \multicolumn{9}{c}{Food} \\
 \hline \noalign{\smallskip}\hline\noalign{\smallskip}

 $\epsilon$ & \multicolumn{9}{c}{1} \\
 
  $t$ & \multicolumn{9}{c}{1} \\

$\lambda$ & \multicolumn{9}{c}{1} \\

 batch size & \multicolumn{9}{c}{4096} \\

learning rate & \multicolumn{9}{c}{0.01} \\

seed  & \multicolumn{9}{c}{0} \\

 $\textit{epochs}_1$ & \multicolumn{9}{c}{200} \\

$\textit{epochs}_2$ & \multicolumn{9}{c}{4000} \\

 \hline

no. of concepts & 8 & 16 & 32 & 64 & 128 & 256 & 1-\textit{pc} & 2-\textit{pc} & full  \\

$m$ & 5 & 5 & 5 & 5 & 5 & 5 & 5 & 5 & - \\

accuracy & 39.10 & 58.58 & 74.42 & 79.91 & 81.61 & 82.59 & 81.86 & 83.01 & 82.55 \\
\bottomrule
\end{tabular}
}
\end{table}

%% file: figures/visualizations.tex
\begin{figure*}[h]
    \centering
    \includegraphics[width=0.7\textwidth]{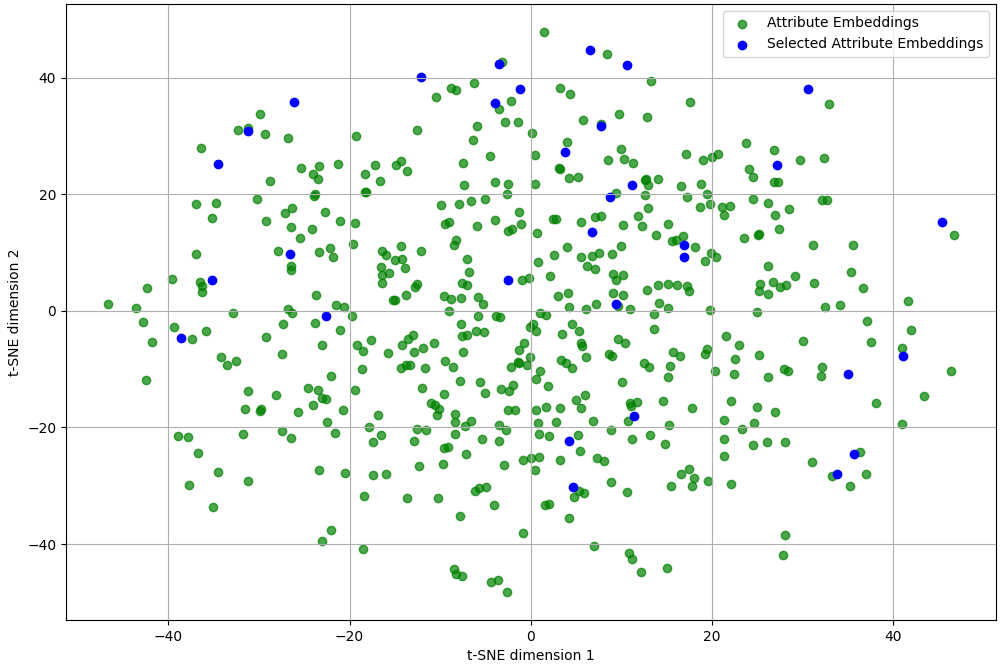}
    \caption{t-SNE visualization of CIFAR-10 descriptors}
    \label{fig:cifar10-atts-vis}
\end{figure*}

\begin{figure*}[h]
    \centering
    \includegraphics[width=0.7\textwidth]{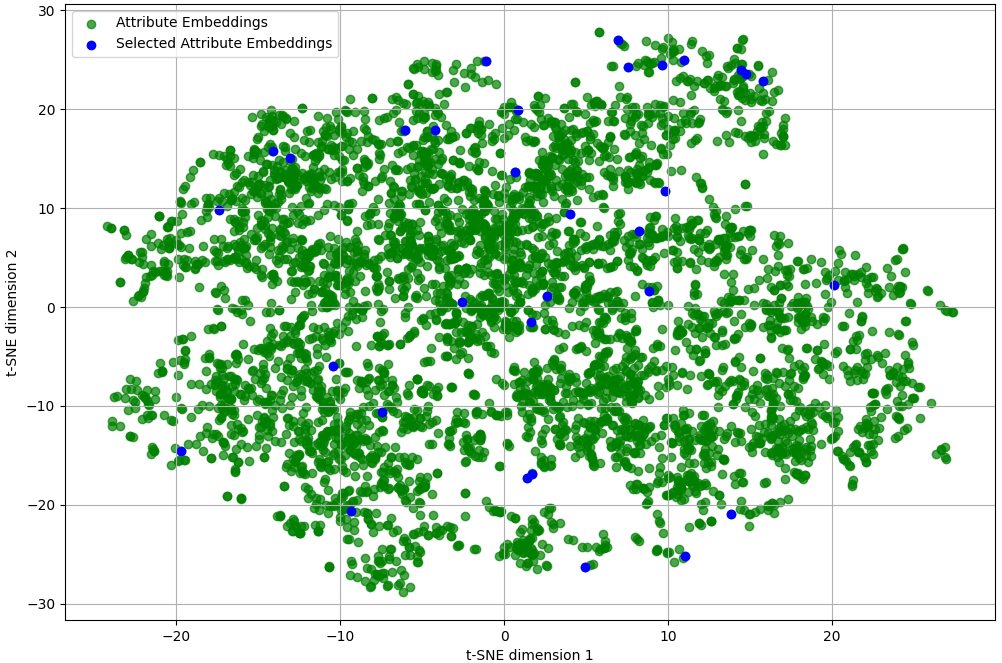}
    \caption{t-SNE visualization of CIFAR-100 descriptors}
    \label{fig:cifar100-atts-vis}
\end{figure*}

\begin{figure*}[h]
    \centering
    \includegraphics[width=0.7\textwidth]{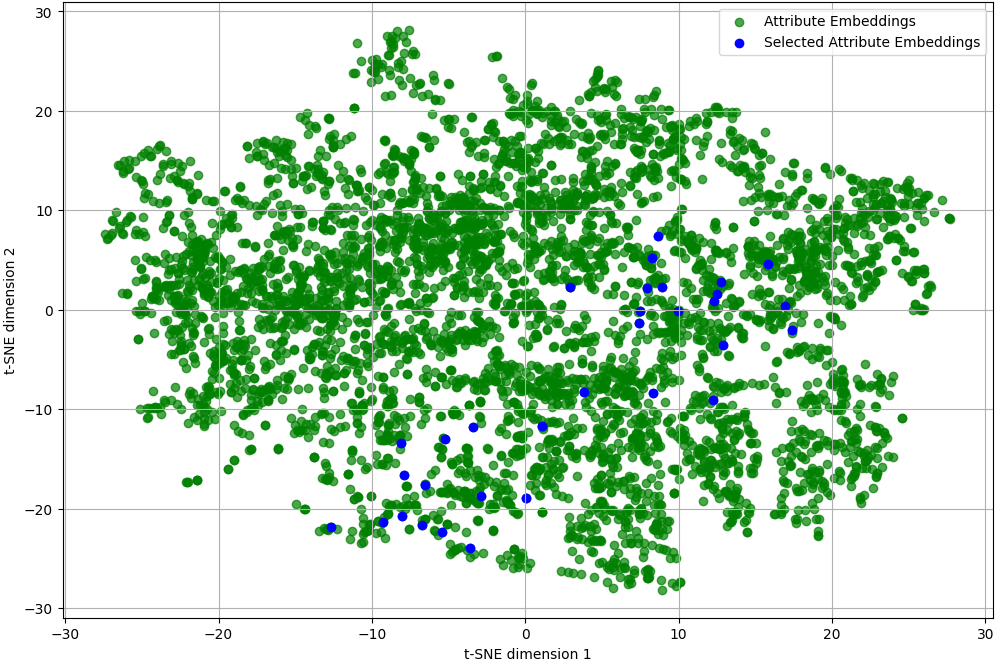}
    \caption{t-SNE visualization of Flower descriptors}
    \label{fig:flower-atts-vis}
\end{figure*}

\begin{figure*}[h]
    \centering
    \includegraphics[width=0.7\textwidth]{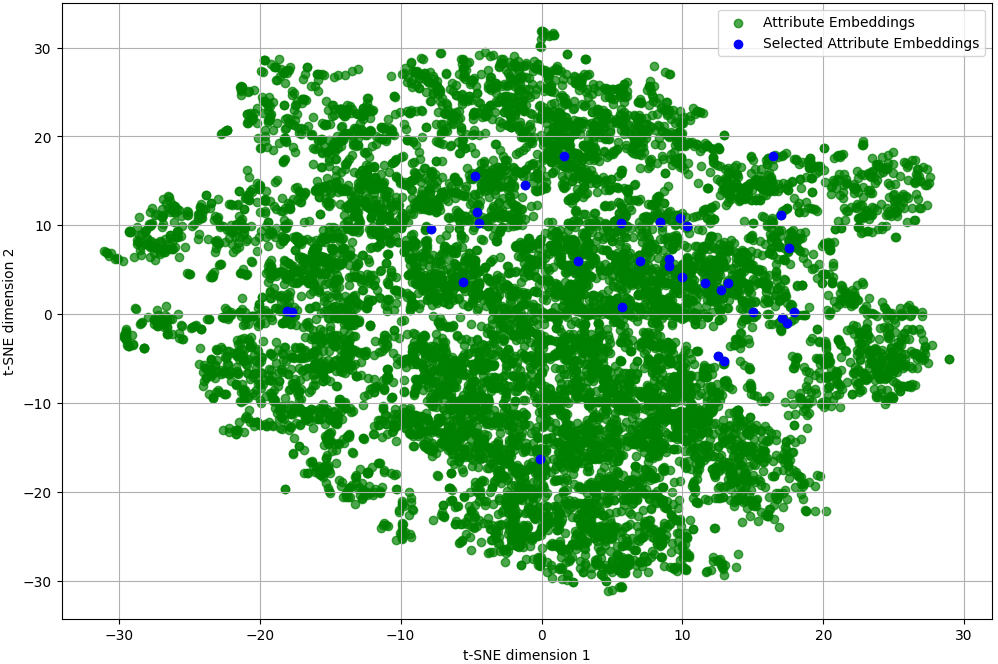}
    \caption{t-SNE visualization of CUB descriptors}
    \label{fig:cub-atts-vis}
\end{figure*}

\begin{figure*}[t!]
    \centering
    \includegraphics[width=0.7\textwidth]{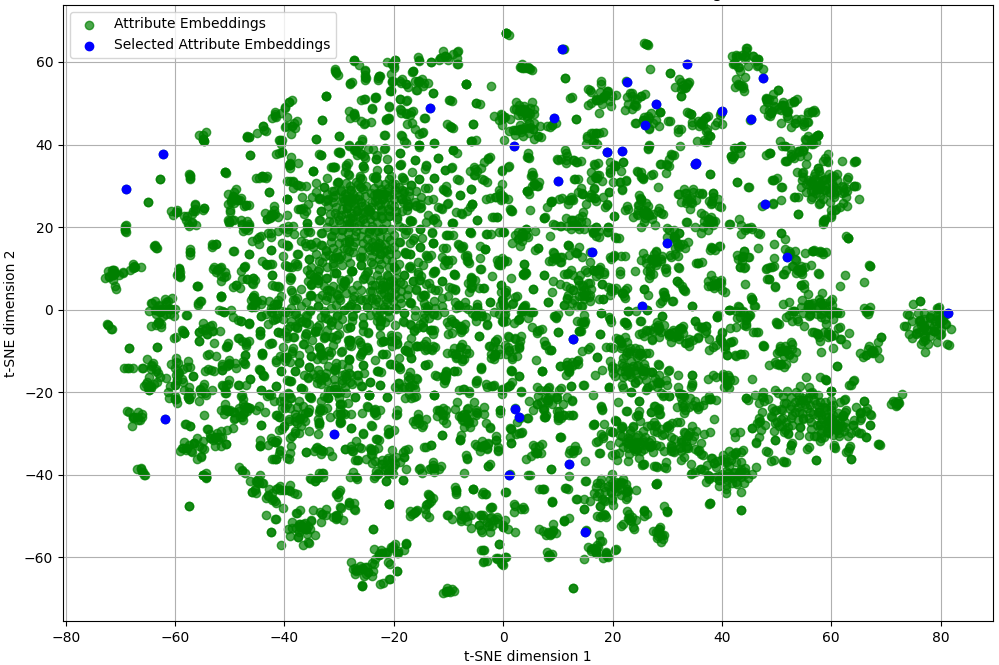}
    \caption{t-SNE visualization of Food descriptors}
    \label{fig:food-atts-vis}
\end{figure*}